\documentclass[letterpaper, 10 pt, conference]{ieeeconf}  % Comment this line out if you need a4paper
\usepackage{lipsum} 
\usepackage{pifont}
\usepackage{amsmath}
\usepackage{amssymb}
\usepackage{mathtools}
\usepackage{stfloats}
\usepackage{subcaption}
\usepackage{float}
\usepackage{tabularx}
\usepackage{booktabs}
\usepackage{multirow}
\usepackage{makecell}

\IEEEoverridecommandlockouts                              % This command is only needed if 
\title{\LARGE \bf
No More Blind Spots: Learning Vision-Based Omnidirectional Bipedal Locomotion for Challenging Terrain
}

\author{Mohitvishnu S. Gadde, Pranay Dugar, Ashish Malik, Alan Fern %
\thanks{*This work is supported by NSF Award 2321851, DARPA contract HR0011-24-9-0423, and the NVIDIA Academic Grant Program.} %
\thanks{All authors are associated with Collaborative Robotics and Intelligent Systems Institute, Oregon State University, Corvallis, OR 97331.}%
\thanks{\{gaddem, dugarp, malikas, alan.fern\}@oregonstate.edu} %
}
\begin{document}

\maketitle
\thispagestyle{empty}
\pagestyle{empty}

%%%%%%%%%%%%%%%%%%%%%%%%%%%%%%%%%%%%%%%%%%%%%%%%%%%%%%%%%%%%%%%%%%%%%%%%%%%%%%%%
\begin{abstract}
Effective bipedal locomotion in dynamic environments, such as cluttered indoor spaces or uneven terrain, requires agile and adaptive movement in all directions. This necessitates omnidirectional terrain sensing and a controller capable of processing such input. We present a learning framework for vision-based omnidirectional bipedal locomotion, enabling seamless movement using depth images. A key challenge is the high computational cost of rendering omnidirectional depth images in simulation, making traditional sim-to-real reinforcement learning (RL) impractical. Our method combines a robust blind controller with a teacher policy that supervises a vision-based student policy, trained on noise-augmented terrain data to avoid rendering costs during RL and ensure robustness. We also introduce a data augmentation technique for supervised student training, accelerating training by up to 10 times compared to conventional methods. Our framework is validated through simulation and real-world tests, demonstrating effective omnidirectional locomotion with minimal reliance on expensive rendering. This is, to the best of our knowledge, the first demonstration of vision-based omnidirectional bipedal locomotion, showcasing its adaptability to diverse terrains.
\end{abstract}

%%%%%%%%%%%%%%%%%%%%%%%%%%%%%%%%%%%%%%%%%%%%%%%%%%%%%%%%%%%%%%%%%%%%%%%%%%%%%%%%
\section{INTRODUCTION}
Legged locomotion has recently been a central objective of the robotics research community due to the capability of such robots to traverse across natural and unstructured terrains \cite{miki_learning_2022,zhuang_humanoid_2024, cheng_extreme_2023, agarwal_legged_2023}. Over the past decade, learning-based approaches, particularly blind locomotion controllers, have demonstrated remarkable progress in enabling stable walking and running across a variety of environments \cite{radosavovic_learning_2024, wu_learn_2025}. These blind policies capitalize on proprioceptive feedback to maintain balance and momentum, forming a strong foundation for real-world deployment \cite{siekmann_blind_2021}. However, despite their robustness on well-modeled or homogeneous surfaces, blind controllers often struggle in the face of unforeseen obstacles or complex, uneven terrains.

To address these limitations, recent research has focused on vision-based locomotion controllers \cite{jenelten_dtc_2024, lee2020learning}, where exteroceptive sensors such as cameras or depth sensors provide crucial information about the environment. By analyzing visual inputs, these controllers adapt to the changes in terrain and proactively adjust foot placements \cite{duan_learning_2022}, thereby improving stability and efficiency. Although such vision-based methods enable new possibilities for agile and adaptive movement, they also introduce challenges in perception, sim-to-real transfer, and high-dimensional policy learning \cite{zhuang_robot_2023}.

% \begin{figure}[!t]
% \centering
% \includegraphics[width=2.5in]{figures/main.png}
% \caption{Place holder fig}
% \label{fig:lead_fig}
% \end{figure}
\setlength{\textfloatsep}{4pt} % Reduce vertical space after float
\setlength{\floatsep}{4pt}     % Optional: reduce space between floats
\setlength{\intextsep}{4pt} 
\begin{figure}[!t]
\centering
\includegraphics[width=0.9\linewidth]{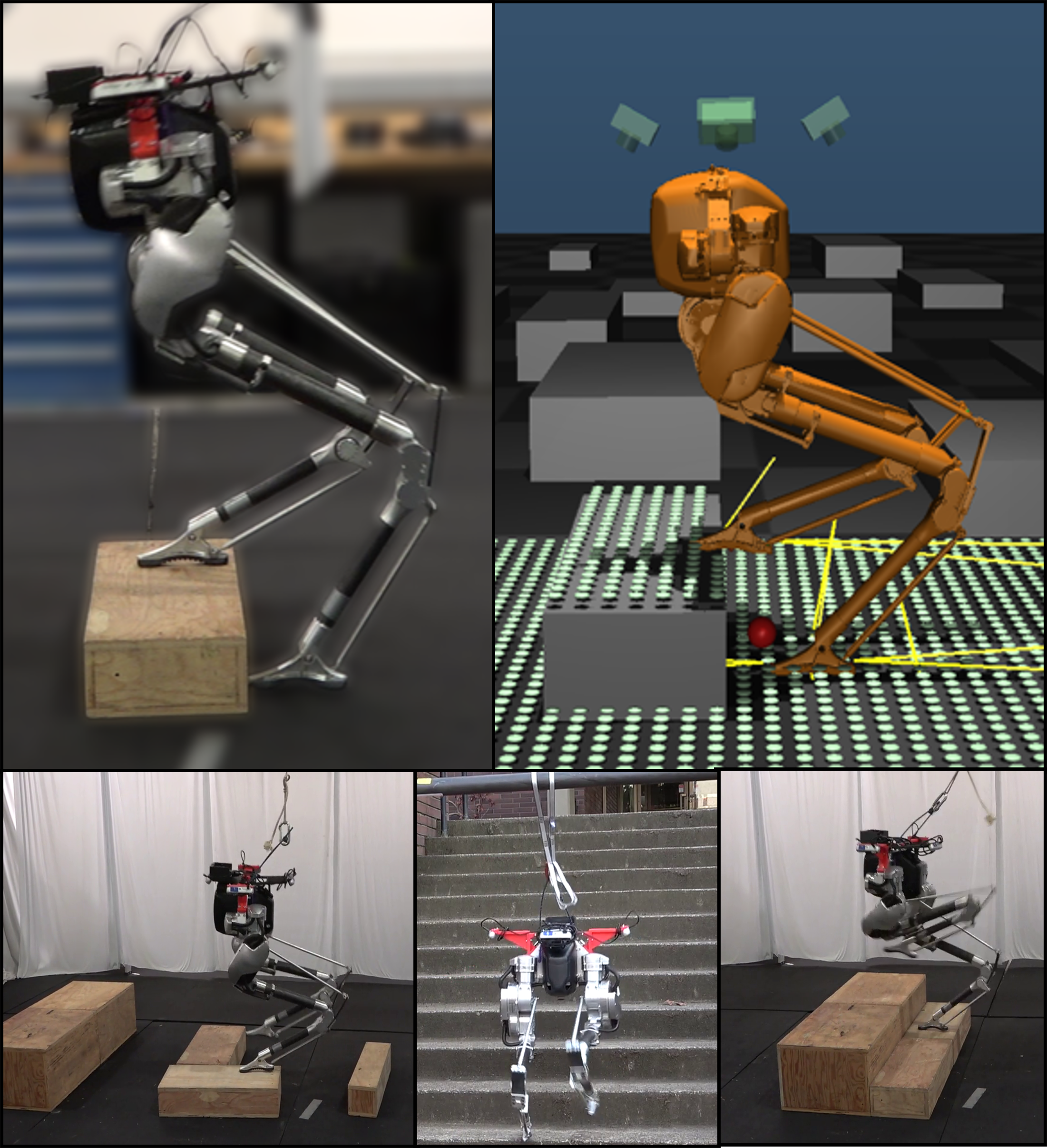}
\caption{The omni-direction robot locomotion controller is trained in simulation via student-teacher learning to proactively adjust the gait to challenging terrain conditions based on the input from 4 depth cameras. This controller is then transferred to the real robot.}
\label{fig:lead_fig}
\end{figure}

Training end-to-end vision-based locomotion controllers is notoriously challenging, particularly when using reinforcement learning (RL), due to the high computational cost of simulating realistic visual inputs. Rendering depth or RGB images for complex environments is expensive \cite{loquercio2022learning}, which slows training and limits throughput. These difficulties are amplified in the omnidirectional setting, where the volume of visual data increases by at least a factor of four, placing even greater demands on simulation and processing. In addition to rendering cost, sim-to-real transfer introduces further complexity: policies must be robust to real-world noise, actuator delays, and the absence of privileged information often used during training. While most prior work on vision-based bipedal locomotion has focused on forward \cite{duan_learning_2024} or constrained directional movement to simplify perception and control, practical applications demand omnidirectional agility, posing a significant challenge for scalable training and deployment of vision-based policies \cite{chen_vmts_2025}.

% Several factors do not allow for seamless design and deployment of robust vision-based locomotion controllers. First, accurately simulating visual perception for rich environments requires very high computational loads, limiting simulation speed and hence througput for training \cite{refs}. Second, policies trained in simulation require significant parameters to be tuned, such as sensor noise and actuator delays,to achieve successful sim-to-real. Third, to train such policies to adapt to varying terrains, significant privileged information is utilized to train such policies that do not allow transfer due to the absence of privileged information in the real world. Finally, end-to-end vision-based policy learning demands significant computational power, careful reward engineering, and high-fidelity rendering, making the development process slow, resource-intensive, and complex. 

% An additional complexity arises from the need for omnidirectional locomotion. Most existing work on vision-based bipedal locomotion (cite helei + couple more) focuses on forward motion or constrained directional control, often limiting their operation in practical applications that demand movement in arbitrary directions. However, this restricted setting helped address the challenge of integrating real-time vision into the high-speed control loop required for dynamic locomotion. Directly extending these prior works to an omnidirectional setting introduces significant computational challenges for both training and testing due to the expanded camera input.  

To address these challenges, we propose a novel framework for vision-based omnidirectional bipedal locomotion that balances computational efficiency with robust real-world performance. The framework is explicitly designed to minimize reliance on computationally expensive depth rendering during training. It achieves this through three key strategies. First, we initialize training with a stable, pre-trained blind locomotion policy that provides basic balance and movement capabilities, serving as a reliable backbone for the vision-based controller. Second, we employ a student-teacher distillation approach enhanced with DAGGER \cite{ross2011}, where a privileged teacher policy is trained via simulation-intensive reinforcement learning using low-cost height maps and environment state information. The vision-based student then learns from this teacher through supervised imitation, using rendered depth images only for the student policy. Third, we introduce a data augmentation technique during student training that significantly increases the effective training set without additional rendering. Together, these components enable efficient training of vision-aware omnidirectional policies within practical time and resource budgets.

% To alleviate these gaps, we propose a novel framework for omnidirectional, vision-aware bipedal locomotion that balances computational efficiency with robust real-world performance. The key challenge is that training bipedal locomotion policies can require a large amount of simulated experience, even for blind locomotion. By naively incorporating expensive visual rendering into the simulation loop to train vision-based locomotion, the computation required for training becomes impractical. We address this complexity in two ways. First, we use a stable pre-trained blind locomotion policy as the backbone of our vision-based locomotion policy architecture, which jump-starts the vision-based training with basic balance and locomotion capabilities. Second, we reduce the amount of expensive rendering during training via student-teacher policy distillation, enhanced with DAGGER (cite) for iterative data aggregation and refinement. This approach first, trains a privileged teacher policy via simulation-intensive reinforcement learning using cheap-to-compute height maps and privileged information about the environment. The vision-based student policy is then trained via supervised imitation learning of the teacher using the more expensive rendered depth inputs. Third, we develop and utilize a data augmentation approach during student-teacher training, which effectively multiplies the amount of imitation training data without requiring additional image rendering.

 Our experimental results demonstrate that our proposed framework leads to significantly faster training than conventional approaches while maintaining competitive performance in both simulation and real-world scenarios. In summary, the key contributions of this work are as follows:
\begin{itemize}%[leftmargin=1em]
\item We present, to the best of our knowledge, the first framework for robust, vision-guided omnidirectional bipedal locomotion.
% \item We integrate a blind locomotion policy as a stable backbone to support both teacher and student training, enabling faster and more reliable policy convergence while reducing reliance on computationally expensive simulations. (not sure about this claim, as one could take it as hierarchical policy or something)
\item We propose an instantiation of student-teacher training specifically targeted at reducing the impractical overhead of directly training with depth rendering in the reinforcement-learning loop. 
\item We propose an innovative data augmentation strategy during student training, where duplicating data buffers and varying velocity commands enhances dataset diversity at a minimal additional computational cost. This approach reduces rendering overhead and shortens overall training time, while improving the policy’s generalization to real-world scenarios.
\end{itemize}

% The remainder of this paper is organized as follows. Section \ref{sec:related} reviews related work in vision-based legged locomotion and teacher-student frameworks. Section \ref{sec:approach} details our problem formulation and presents the proposed method. In Section \ref{sec:results}, we provide extensive simulation results, including ablation studies. Section \ref{sec:hardware} discusses the deployment of the method on hardware. Finally, Section \ref{sec:summary} addresses the current limitations and future directions.

\section{Related Work}
\label{sec:realated}

\subsection{Reinforcement Learning (RL) for Bipedal Robots}
In recent years, RL-based controllers have been extensively researched for legged locomotion \cite{siekmann_blind_2021, vanMarum2024, radosavovic_learning_2024}. These controllers are usually trained in simulation using model-free RL approaches, which leverage the robot's proprioception, enabling legged robots to traverse various terrains without taking any explicit gait input to achieve locomotion. However, despite the advances, these blind locomotion policies primarily rely upon the proprioceptive feedback from the robot to learn a policy that is able to achieve stability and maintain balance during locomotion in the presence of moderate disturbances. This allows for locomotion on well-structured surfaces with limited amounts of ground and external disturbances. 
%However, the operation of such controllers is limited when encountering highly dynamic and unpredictable terrains. 

Unfortunately, blind RL-based locomotion controllers struggle with unknown obstacles and significant terrain variations as they cannot sense or anticipate the changes in the environment they are operating under before making foot placements. This limitation often results in suboptimal locomotion performance and requires excessive corrective inputs to traverse such environments to avoid failures. Vision-based locomotion has been explored as a means to provide exteroceptive sensing to address these challenges, allowing these legged robots to perceive and interpret the terrain features and traverse varying terrain.
.

\subsection{Vision-Based Legged Locomotion}

%Vision-based locomotion has been extensively researched for quadrupedal robots (cite). These methods typically use RL methods to train and demonstrate sim-to-real transfer of these vision-based locomotion controllers. Elevation maps or height scans in the global frame of reference have been used as the observed perceptive input to train the RL policy. For example, bipedal robots(cite) and quadrupedal robots(cite) utilize these height map scans around each foot. While several other approaches (cite) employ a uniformly structured elevation map that is centered at the robot frame. Thus, making these approaches highly sensitive to sensor calibration. In contrast, in this work, we
%are interested in a solution that does not require careful
%calibration for global odometry.  Some other types of approaches
% directly use vision inputs from cameras, such as depth
% images [cite] or RGB images [cite], which are used as inputs to an RL policy. This end-to-end training is often carried
% out via teacher-student training (cite), where the teacher model benefits from the privileged information in simulation to enhance policy learning. These methods have demonstrated success in deploying these RL-trained policies on quadrupedal robots in real-world scenarios and present significant potential for application in bipedal locomotion.

Vision-based locomotion has been extensively researched for quadrupedal robots \cite{yu_visual-locomotion_nodate, miki_learning_2022,cheng_extreme_2023, agarwal_legged_2023} and, to a lesser extent, for bipedal robots \cite{duan_learning_2024, chen_vmts_2025}. These methods typically use RL to train and demonstrate sim-to-real transfer of these vision-based locomotion controllers. Elevation maps or height scans in the global frame of reference have been used as the observed perceptive input to train the RL policy. For example, bipedal robots\cite{duan_learning_2024} and quadrupedal robots\cite{fankhauser_perceptive_2018, margolis_learning_2021, hoeller_neural_2022, miki_elevation_2022} utilize these height map scans around each foot. While several other approaches \cite{fankhauser_perceptive_2018, margolis_learning_2021} employ a uniformly structured elevation map that is centered at the robot frame. By assuming the availability of height maps at test time, simulation-based training is able to avoid expensive rendering of image and depth data, making training via RL practical. However, this assumption requires careful sensor calibration and robust height map estimation to maintain consistency between the simulated experience and real-world. 

In contrast, we are interested in a solution that does not require careful calibration for global odometry or direct estimation of local elevation maps. This is motivated by the additional complexity of such calibration and estimation in the omnidirectional setting, where multiple cameras are typically involved (4 in our experiments). Prior work has considered learning quadruped locomotion policies that directly use vision input from cameras, including depth images \cite{hoeller_neural_2022, miki_elevation_2022} and RGB images \cite{loquercio_learning_2023}. This type of end-to-end training is often carried out via teacher-student approaches \cite{wang_cts_2024, yu_visual-locomotion_nodate, miki_learning_2022,cheng_extreme_2023, agarwal_legged_2023}, where the teacher model benefits from the privileged information in the simulation to enhance policy learning. These methods have demonstrated success in deploying these RL-trained policies on quadrupedal robots in real-world scenarios. However, they have yet to be demonstrated on the more difficult problem of bipedal locomotion. Our work expands on these end-to-end student-teacher approaches to make them practical for the more computationally intensive training of bipedal locomotion. 

% \subsection{Teacher Student Approaches}
% Teacher-student learning frameworks have been introduced to bridge the gap between privileged training information in simulations and realistic sensor data available in real-world deployment. These frameworks first train a privileged "teacher" policy with access to complete environmental information and then distill the knowledge into a "student" policy that only relies on realistic sensory inputs. Several approaches, including domain adaptation (cite), imitation learning (cite), and progressive policy refinement (cite), have been explored to enhance the efficiency and robustness of the student policy.

\begin{figure*}[!htb]
    \centering
    % \vspace{5pt}
    \includegraphics[width=0.85\linewidth]{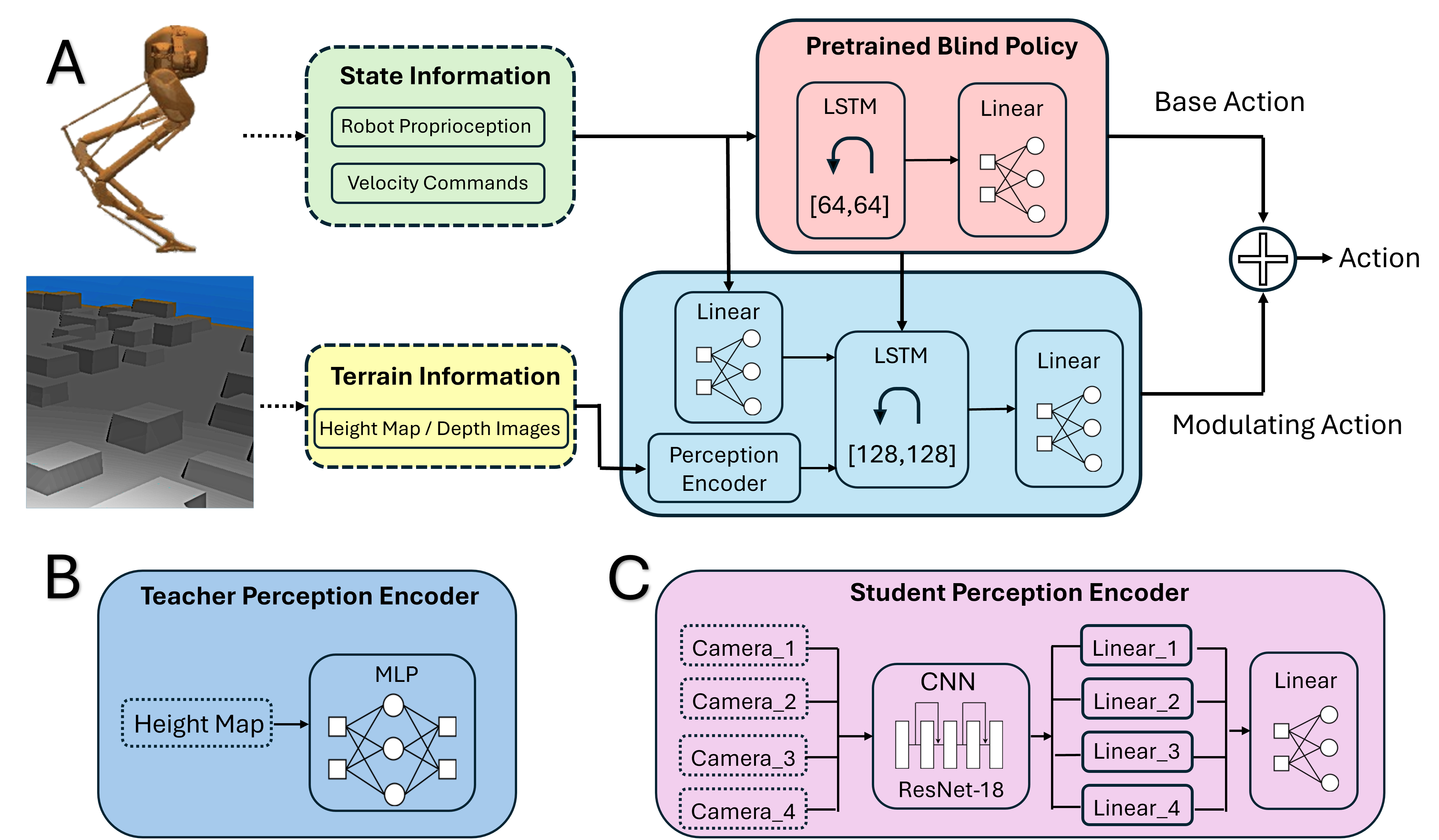}
    \caption{ Hierarchical network architecture for proposed vision-based omnidirectional locomotion control. \textbf{(A)} The policy consists of a frozen pretrained blind policy that outputs a base action, and a trainable modulator conditioned on perception to produce a modulating action. The final action is the sum of both outputs. \textbf{(B)} The teacher model uses privileged height map encoded by an MLP. \textbf{(C)} The student model uses egocentric multi-view images processed by a ResNet-18-based encoder.}
    % \vspace{-4pt}
    \label{fig:hierarchical-architecture}
    \vspace{-2em}
\end{figure*}

\section{Training an Omnidirectional Terrain-Aware Locomotion Policy}
\label{sec:approach}
We formulate the omnidirectional terrain-aware locomotion problem as a Markov Decision Process (MDP), defined by the tuple \(\mathcal{M} = (\mathcal{S}, \mathcal{A}, T, r, \gamma)\), where the state space \(\mathcal{S}\) consists of the robot’s proprioceptive and exteroceptive observations, including joint positions, velocities, base orientation, linear and angular velocities, and terrain-related sensory inputs such as height maps or depth inputs. The action space \(\mathcal{A}\) represents low-level motor control commands. \(T\) represents the system transition function. The agent receives a reward \(r(s_t, a_t)\) at each time step to encourage stable and efficient locomotion, with reward terms designed to promote following commanded velocity, minimizing energy consumption, maintaining balance, and successfully traversing terrains. The objective is to learn a policy \(\pi(a_t | s_t)\) that maximizes the expected cumulative reward \(J(\pi) = \mathbb{E} \left[ \sum_{t=0}^{T} \gamma^t r(s_t, a_t) \right]\), where \(\gamma\) is the discount factor balancing immediate and future rewards. 

\subsection{Approach Overview}

Training bipedal locomotion controllers via RL requires extensive simulation experience. While this has been practical for blind locomotion, the computational cost of simulation becomes impractical when visual rendering must be done in the RL loop. Thus, our approach is designed to minimize the amount of rendering done in simulation during training. In particular, we enhance the popular student-teacher framework that has been used successfully in prior robotics work, including quadruped locomotion \cite{wang_cts_2024, yu_visual-locomotion_nodate, miki_learning_2022,cheng_extreme_2023, agarwal_legged_2023, zhang_resilient_2024}, manipulation \cite{yamada_twist_2023, hu_privileged_2023, chen_system_2022, pinto_asymmetric_2017}, and self-driving \cite{chen_learning_2020, bansal_chauffeurnet_2018, bewley_learning_2018}. 

The general student-teacher approach first trains a teacher policy via simulation-intensive RL using privileged information, which is typically intended to decrease the difficulty of learning. Next, a student policy is trained to mimic the teacher policy without using privileged information, but rather input information that will be available at test time. Importantly, student training is fundamentally a supervised learning problem, which avoids the simulation-intensive exploratory learning process of RL. 

In this work, we instantiate the student-teacher framework to avoid rendering during teacher training. In particular, the teacher has access to easily available height map, which requires minimal computational overhead. It is only during student training that depth map rendering is included in the simulation loop. Our approach additionally enhances the student-teacher framework in two ways to further improve the cost of training. First, we use a hierarchical policy architecture that leverages a pre-trained blind policy for both the student and teacher. Second, we develop a data augmentation approach during student training that effectively multiplies the amount of training data without additional rendering or simulations. Below, we describe the details of the policy architecture and the student-teacher training approach. 

\subsection{Policy Architectures}

% The student and teacher use nearly identical policy architectures, differing only in the input blocks corresponding to visual perception. Below, we describe the observation space, the common architecture that assumes a provided visual-input encoding. Next, we describe the specific visual input encoders for the privileged teacher and student. 
The student and teacher policies share a nearly identical neural network architecture, differing only in the input blocks used for visual perception. Below, we first describe the observation space and the shared policy architecture, which assumes a generic visual input encoding. We then detail the specific vision encoders used by the teacher and student, corresponding to privileged height maps and egocentric depth maps, respectively.

\subsubsection{Observation and Action Spaces}
% ......  describe the generic architecture in terms of a generic visual input encoding (these will be produced by the teacher and student vision blocks). You would clarify the inputs and outputs of the policy and then give the architecture. 

Both student and teacher policies are trained to operate using a combination of the robot's proprioceptive state, perceptual representation of surrounding terrain, and user-specified control commands, which are described as follows: 
\begin{itemize}%[leftmargin=1em]
    \item \emph{Robot proprioceptive state:} This includes the position and velocity of all the measurable actuated and un-actuated joints, along with the orientation (in quaternion) and angular velocity of the floating base.
    \item \emph{Perceptive input:} This includes information about the local terrain around the robot. The teacher policy uses a privileged height map centered at the robot, while the student policy receives egocentric depth images captured in real time from four cameras covering rectangular regions around the robot.   
    \item \emph{User control commands}: These define the robot's target motion and consist of desired linear velocities in the X and Y directions, and angular velocity about the yaw axis.
\end{itemize}

The action space for both teacher and student policies are the PD position setpoints for all 10 actuators (5 per leg). These setpoints are produced by our policy at 50Hz and processed by low-level PD controllers running at 2 kHz. This setup ensures smooth, responsive execution of the high-level actions produced by the network, enabling stable and adaptive locomotion across diverse terrains.

\subsubsection{Network Architecture}

Both policies follow a hybrid neural network architecture that combines the action produced by a pre-trained blind locomotion controller with a differential action computed by a network that depends on the perceptual input. This choice of architecture enables both policies to leverage stable low-level locomotion skills while adapting to complex terrains through visual feedback. Intuitively, on moderate terrain the differential action can be nearly zero, since the blind policy is already effective, and need only adjust the blind policy action as necessary based on the visual terrain input. The policy architecture is shown in Figure \ref{fig:hierarchical-architecture} and includes the following components. 
\begin{itemize}%[leftmargin=1em]
    \item \emph{Pre-Trained Blind Locomotion Controller}: The blind locomotion policy is adapted from \cite{vanMarum2024}, which is a 2-layered LSTM network with 64 hidden units designed to learn robust locomotion using proprioceptive feedback. This policy is trained on relatively flat terrain to follow linear and angular velocity commands. . To enhance stability and robustness, the training process incorporates external perturbations, allowing the robot to maintain stability and recover from disturbances during locomotion. This policy is pre-trained and remains frozen in our architecture during omnidirectional training. 
    \item \emph{Proprioceptive and Command Encoder}: The robot’s joint positions and velocities, floating base states, and user-specified control commands are encoded using a 64-dimensional linear layer into a latent feature vector.
    \item \emph{Teacher Vision Encoder}: The teacher policy processes the four concatenated privileged height maps via a two-layer MLP with [256, 128] units per layer and ReLU activation functions resulting in a 128-dimensional perceptual embedding. 
    \item \emph{Student Vision Encoder}: The student policy's vision encoder is depicted in Figure \ref{fig:hierarchical-architecture}-(C) . It first uses a shared parameter ResNet-18 convolutional network to independently encode each depth images captured from the four cameras. The encodings of each image are then projected into a 128-dimensional encoding using individual single-layer MLPs. Finally, all four encodings are concatenated and passed through a final single-layer MLP to get a single 128-dimensional perceptual embedding.
    \item \emph{Differential Action Network}: The encoded proprioceptive-command vector, the perceptual embedding, and the action output from the blind locomotion controller are concatenated to form a unified feature representation. This combined representation is passed through a two-layer LSTM network, each layer consisting of 128 hidden units. The LSTM captures temporal dependencies across observation sequences and enables the policy to adapt its behavior based on both proprioceptive and visual feedback over time. The output of the LSTM is a differential action vector with the same dimensionality as the action space.
    \item \emph{Action Output}: The final action is computed by adding the differential action to the baseline action from the blind locomotion controller. 
\end{itemize}
% \subsubsection{Action Space}: Both the teacher and student policies operate at a control frequency of 50Hz and output PD position setpoints for all 10 actuators. These setpoints are generated by the policy network, as described above, and are provided as target positions to a low-level PD controller running at 2 kHz. This setup ensures smooth, responsive execution of the high-level actions produced by the network, enabling stable and adaptive locomotion across diverse terrains.

\subsection{Privileged Teacher Training}
The teacher policy is trained using the Proximal Policy Optimization (PPO) actor-critic RL algorithm \cite{schulman2017proximal} in a MuJoCo simulation environment \cite{todorov2012mujoco}, where the robot aims to follow randomized commands over diverse and randomized terrains. The training is conducted using 120 cores on a dual Intel Xeon Platinum 8280 server on the Intel vLab Cluster. Below, we describe our training environment and the training episode generation, followed by our choice of reward function, and finally details of our PPO implementation. 

\begin{figure}[!h]
\centering
\includegraphics[width=\linewidth]{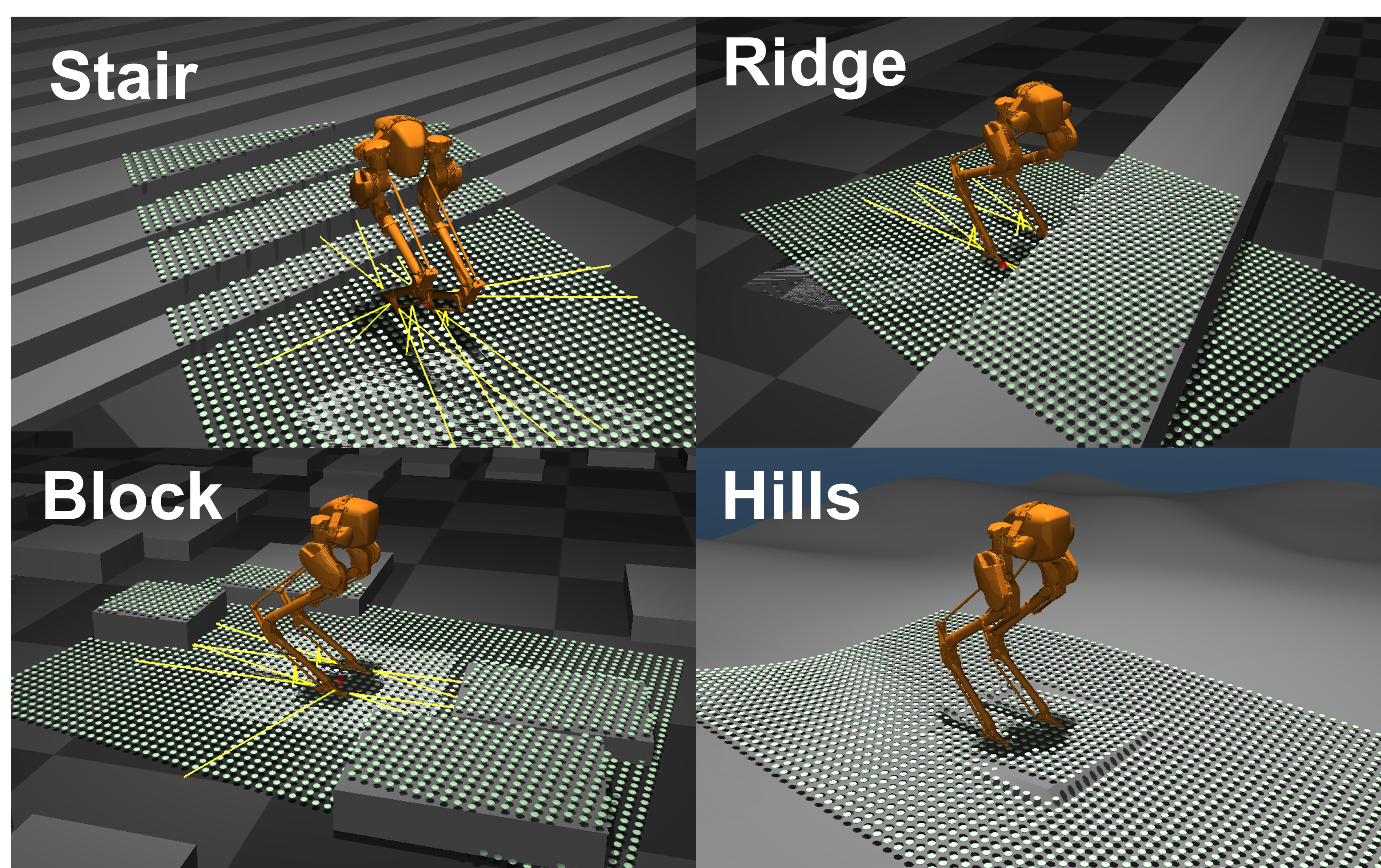}
\caption{Various types of terrains used in training}
\label{fig:terrain}
\end{figure}

\noindent {\bf Training Episode Generation.} 
Each training episode takes place on a randomly generated 30\,m $\times$ 30\,m terrain map, drawn from one of five terrain types illustrated in Figure~\ref{fig:terrain}: 
(1) \emph{Flat} – featureless terrain, 
(2) \emph{Hills} – smoothly varying height maps, 
(3) \emph{Blocks} – randomly placed blocks of varying dimensions, 
(4) \emph{Ridges} – sequential ridges with varying width and height, and 
(5) \emph{Stairs} – upward and downward stairs with varying step sizes. 
These terrain types are sampled with probabilities [0.03, 0.07, 0.35, 0.20, 0.35], respectively. 
Randomization parameters are detailed in Table~\ref{tab:terrain-rand}, with care taken to avoid feature overlap.
\setlength{\textfloatsep}{4pt} % Reduce vertical space after float
\setlength{\floatsep}{4pt}     % Optional: reduce space between floats
\setlength{\intextsep}{4pt} 
\begin{table}[!h]
\centering
% \scriptsize
\setlength{\tabcolsep}{3pt}
\renewcommand{\arraystretch}{1.15}
\caption{Ranges for terrain randomization used in training and evaluation. All terrains are uniformly sampled within the specified range.}
\label{tab:terrain-rand}
\begin{tabular}{|l|l|c|c|c|}
\hline
\textbf{Terrain} & \textbf{Parameter} & \textbf{Train Range} & \textbf{Easy Eval} & \textbf{Hard Eval} \\
\hline
Ridge & Height [m] & [0.05, 0.6] & [0.05, 0.5] & [0.5, 0.6] \\
\hline
\multirow{3}{*}{Stair} 
& Height [m] & [0.05, 0.2] & [0.05, 0.2] & [0.2, 0.3] \\
& Length [m] & [0.25, 0.4] & [0.4, 0.4] & [0.25, 0.4] \\
& Steps      & [4, 28]     & [4, 12]     & [12, 28] \\
\hline
\multirow{2}{*}{Block}
& Len./Width [m] & [0.4, 1.0] & [1.0, 1.0] & [0.4, 1.0] \\
& Height [m]     & [0.05, 0.4] & [0.05, 0.3] & [0.3, 0.5] \\
\hline
\end{tabular}
\end{table}

A small portion of episodes use flat and hilly terrains to stabilize training, while the majority focus on blocks, ridges, and stairs, which require more complex integration of visual and proprioceptive feedback. We observe that terrain diversity is more effective than reward tuning for encouraging robust, non-aggressive behaviors. For instance, omitting easier terrains leads to overly aggressive gaits, while incorporating realistic variations in stair step length or ridge height encourages more natural step regulation through perception.

The height map provided to the robot at each time step is a 3\,m $\times$ 2\,m area centered on the robot. The map is rendered at 5\,cm resolution and encodes ground height relative to the robot’s floating base. This relative encoding supports terrain perception without requiring global localization or odometry, which are often unreliable in real-world settings.

At the start of each episode, the robot is placed randomly near the center of the terrain in a standing pose, facing a random direction. It is then given a randomly sampled velocity command, drawn from the set: step-in-place, step-in-place-turn, walk, and walk-turn, with probabilities [0.05, 0.05, 0.60, 0.30]. A corresponding height command is also provided. Velocity commands are sampled uniformly from the following ranges: 
X velocity $\in [-0.6, 1.0]$\,m/s, 
Y velocity $\in [-0.45, 0.45]$\,m/s, 
yaw rate $\in [-22.5, 22.5]$\,deg/s, 
and base height $\in [0.4, 0.95]$\,m. 
To promote adaptation, the command is randomly modified once during the episode at a timestep between 200 and 250.

Each episode runs for up to 450 timesteps (9 seconds simulated). Early termination occurs if any of the following conditions are met: (1) The floating base roll or pitch exceeds 15 degrees, (2) The robot’s linear velocity norm exceeds 1 plus the commanded velocity, (3) The base height drops below 40\,cm relative to terrain, and (4) The robot’s body (excluding conrods and feet) collides with the terrain.

\noindent \textbf{Reward Function.} Our reward function is designed to promote smooth, stable, and transferable bipedal locomotion over complex terrain. It encourages clean contact transitions, discourages unsafe configurations, and supports robust gait formation. The total reward is the sum of three equally weighted components:
\[
R = R_{0} + R_{\text{feet}} + R_{\text{con}}
\]

\begin{itemize}%[leftmargin=1em]
    \item \textit{Base Locomotion Reward ($R_0$):} Adapted from \cite{vanMarum2024}, this reward stabilizes a consistent alternating gait and forms the foundation of both the blind controller and the vision-based policy. It promotes agreement with target commands, balance, and regular stepping.
    \item \textit{Foot Collision Penalty ($R_{\text{feet}}$):} This term penalizes unwanted collisions between the front tip of the foot and the terrain during swing. A penalty of $-5$ is applied for each such contact, which helps the policy learn to lift the feet cleanly. Without this term, agents tend to stumble, harming sim-to-real transfer. In simulation, we detect these events using contact sensors at the front of each foot. These sensors are used only during training for reward computation and are not required at test time.
    \item \textit{Conrod Contact Penalty ($R_{\text{con}}$):} To avoid unsafe leg configurations, especially during sideways or backward locomotion. This term penalizes contact between the robot’s conrods and the terrain. It helps the policy learn to maintain safe leg posture and avoid dragging or scraping parts that could cause instability or physical damage.
\end{itemize}

\noindent \textbf{PPO Modifications.} To improve training stability, we introduce two modifications to the standard PPO algorithm. First, we modify the PPO loss to include a mirror loss over robot proprioceptive inputs as well as visual inputs. This loss encourages the policy to choose symmetric actions when facing terrain that is symmetrically mirrored about the sagittal plane. Second, since our reward function uses privileged information, we enhance the critic network by providing it with additional privileged inputs beyond those available to the actor policy. These include the robot’s height and foot positions in the global frame, a square 0.5m height map around each foot, and multiple rangefinder sensors on both legs, two in front of each foot, two on the sides, two at the rear, and two on the conrods. These range measurements are used only by the critic to enhance terrain understanding around contact regions. These additional inputs improve the critic's ability to predict long-term reward, particularly for complex terrain interactions that are not fully observable from proprioception or height maps alone.
\vspace{-0.7em}

% \begin{figure}[!h]
% \centering
% \includegraphics[width=\linewidth]{figures/depth_v2.png}
% \caption{depth image noise addition}
% \label{fig:noise_depth}
% \end{figure}

\subsection{Student Training}
The training process for the student policy largely mirrors that of the teacher policy, including terrain generation, sampling of velocity and height commands, and episode termination conditions. However, unlike the teacher, which is trained via RL, the student policy is trained through policy distillation. Specifically, we follow the DAGGER training strategy \cite{ross2011}, where each episode is generated by using the student (initially random) to select actions with the privileged teacher being applied at each step to generate supervisory training data. The student is trained via random mini-batches collected across the episodes to mimic the teacher's actions and perceptual encodings. In particular, the student's loss function includes two terms: 1) the mean squared error (MSE) between the teacher and student action, and 2) the MSE between the teacher's perceptual encoding and the student's encoding. We found that the second term helps accelerate learning by directly supervising the ResNet image encoder to produce a representation that the teacher discovered during privileged training. 
%This loss is calculated over both the predicted action distributions and the internal perceptual embeddings, encouraging the student to match not only the outputs but also the underlying feature representations.

%The key architectural difference lies in the perception module: the teacher uses a privileged height map encoder, while the student replaces this with a ResNet-18-based CNN to process real-time depth images, as described in the previous sub-section.

Note that since the student requires depth renderings for its input, generating these trajectories is computationally expensive compared to teacher training. To make more efficient use of this data and improve generalization, we introduce a targeted data augmentation strategy. Rather than collecting more simulation episodes, we generate additional supervision signals by duplicating existing state trajectories and varying the velocity commands given to the teacher at each time step. Thus, a single rendered depth image at a timestep results in multiple training examples. This imputation technique simulates how the teacher would respond to different locomotion intents in the same terrain context.

Concretely, for each training trajectory buffer, we sample random values within bounded ranges for X velocity, Y velocity, and yaw rate, drawn from: \emph{X velocity:} [-0.6, 1.0] m/s, \emph{Y velocity:} [-0.6, 0.6] m/s, and \emph{Turn rate: }[-0.4, 0.4] rad/s.
These new command values are injected into both the student and teacher input states at the corresponding command indices. The teacher and student policies are then evaluated on these modified inputs. The student is trained to minimize the KL divergence between the resulting actions under these perturbed commands. This process is repeated over several epochs (e.g., 10 per batch), increasing training diversity without additional simulation steps.

To further improve robustness to sensor noise and real-world imperfections, we apply standard vision augmentations to the student’s input depth images. These include Gaussian noise injection, dropout artifacts, and random image jittering; simulating the variability and degradation seen in real camera data. Together, these modifications ensure that the student policy faithfully transfers the terrain-adaptive locomotion strategy learned by the teacher to a real-world deployable form, relying solely on realistic depth-based perception without access to privileged simulation data.

\setlength{\textfloatsep}{4pt} % Reduce vertical space after float
\setlength{\floatsep}{4pt}     % Optional: reduce space between floats
\setlength{\intextsep}{4pt} 

\begin{table}[!t]
\centering
\caption{Parameters and ranges used in domain randomization. All values are uniformly sampled within the range.}
\label{tab:policy-dr}
\renewcommand{\arraystretch}{1.2}
\setlength{\tabcolsep}{4pt}
\begin{tabular}{|c|c|c|c|}
\hline
\multicolumn{2}{|c|}{\textbf{Parameter}} & \textbf{Range} & \textbf{Unit} \\
\hline
\multirow{6}{*}{Sim Model} 
% & Joint Damping      & [0.5, 2.5]            & \%     \\ \cline{2-4}
& Mass               & [-0.25, 0.25]         & \%     \\ %\cline{2-4}
& COM Location       & [-0.01, 0.01]         & m      \\ %\cline{2-4}
& Spring Stiffness   & [-500, 500]           & Nm/rad \\ %\cline{2-4}
& Torque Efficiency  & [0.9, 1.0]            & \%     \\ %\cline{2-4}
& Torque Delay       & [0.5, 3]              & ms     \\ %\cline{2-4}
& Encoder Noise      & [-0.05, 0.05]         & rad     \\
\hline
\multirow{4}{*}{Height Map} 
& XY Shift / (episode,step) & [-0.05, 0.05] & m    \\ %\cline{2-4}
& Z Shift / episode  & [-0.1, 0.1]  & m \\ %\cline{2-4}
& Z Shift / step     & [-0.02, 0.02]   & m \\
& Delay              & [20, 100]             & ms     \\
\hline
\end{tabular}
\end{table}
\subsection{Domain Randomization}

To ensure robust sim-to-real transfer and to increase training data diversity, we apply extensive domain randomization across model parameters, actuation characteristics, sensor inputs, and actuation delays. A summary of the randomized parameters is shown in Table \ref{tab:policy-dr}.

Model parameters such as mass and link inertia are randomized per episode to simulate variability across different robots. Randomizing torque efficiency is particularly impactful for transfer to extreme terrains (e.g., stepping up 0.5m), where torque saturation of the knee motor can otherwise limit performance. Actuation delays are also introduced by randomly delaying torque commands by up to 3 milliseconds.

Visual input perturbations are introduced to prevent the policy from overfitting to simulation-generated terrain representations. The height map is spatially shifted per episode and at every policy step to simulate temporal noise, and a randomized delay of up to 100 milliseconds is applied to simulate perception latency. These perturbations are critical for preparing the teacher policy to operate under imperfect and noisy conditions that the student will encounter.

\section{Simulation Experimental Results}
\label{sec:sim-results}

\subsection{Policy Performance}
We evaluate the performance of the state-of-the-art blind bipedal locomotion controller, SaW \cite{vanMarum2024}, alongside three policies trained using our proposed approach: (1) a blind policy trained from scratch on the same terrain distribution described in the previous sections, (2) the proposed full privileged teacher policy, and (3) the distilled student policy that operates with egocentric depth inputs. %This comparison highlights the impact of visual perception and policy architecture on locomotion robustness and adaptability across diverse terrains.

\begin{figure*}[!htb]
    \centering
    \vspace{5pt}
    \includegraphics[width=\linewidth]{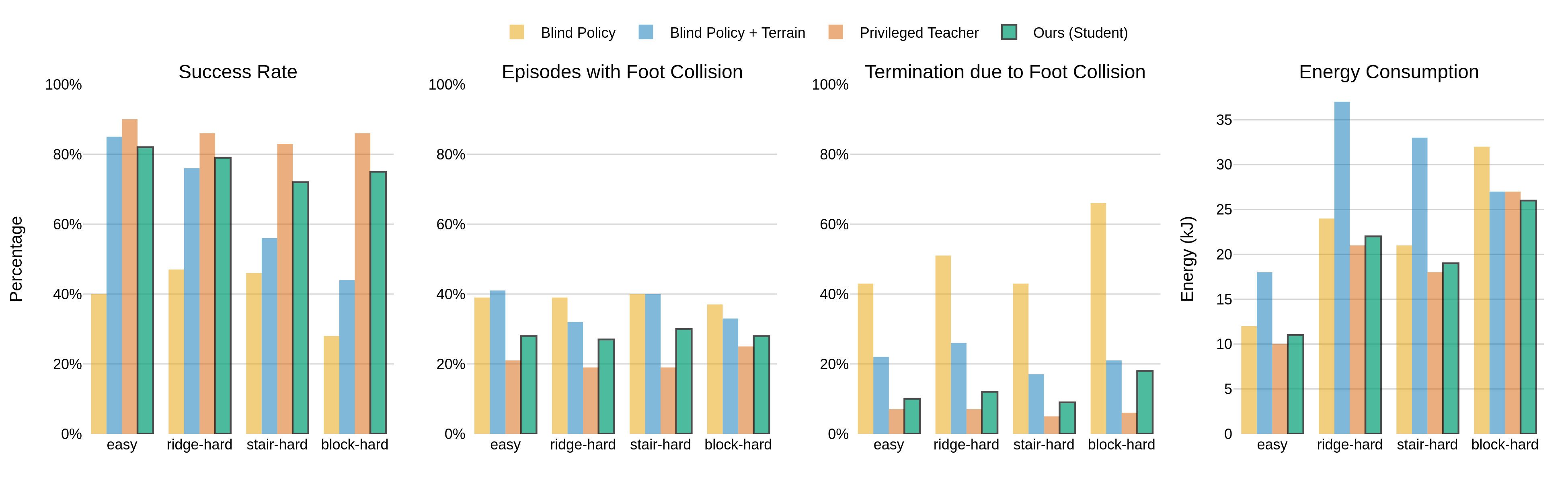}
    \caption{Bar plots comparing four policy variants: \emph{Blind Policy, Blind Policy + Terrain, Privileged Teacher, and Ours (Student)} across four terrain types (easy, ridge-hard, stair-hard, block-hard). The plots show (1) Success Rate, defined as completing 10-second rollouts without falling; (2) Episodes with Collision, which count any episode with foot or conrod collision events; (3) Terminations due to Foot Collision, measured by checking if foot or conrod collisions occurred in the final 100 steps of terminated episodes; and (4) Energy Consumption, reported as average cumulative joint effort in kilojoules. All metrics are averaged over 100 episodes for each variant of terrain with 95\% confidence intervals.}
    \label{fig:evaluation}
    \vspace{-2em}
\end{figure*}

To understand the contribution of each policy component, we conduct an ablation study in which each terrain is evaluated under two difficulty levels: easy and hard, as defined in Table \ref{tab:terrain-rand}. For each policy and terrain mode, we collect 100 episodes and compute four key evaluation metrics: \emph{Success Rate, Episodes with Foot Collisions, Terminations due to Foot Collisions}, and \emph{Energy Consumption}, as shown in Fig. \ref{fig:evaluation}. Note that for space constraint reasons, results for ``easy" are aggregated across the three terrain type, while the ``hard" category is broken down by terrain.  

\setlength{\textfloatsep}{4pt} % Reduce vertical space after float
\setlength{\floatsep}{4pt}     % Optional: reduce space between floats
\setlength{\intextsep}{4pt} 
In terms of Success Rate, the Privileged Teacher consistently outperforms all baselines across terrains, achieving over 85\% success even on the most difficult terrains. The vast majority of failures occur during episodes that require high sideways or backward stepping. 
The robot morphology, particularly the conrods, makes such high steps extremely difficult. It is unclear how close the Teacher's performance is to the maximum achievable on this terrain distribution given the robot morphology. The nonprivileged vision-based student policy closely matches the teacher's performance, notably outperforming both blind variants, especially on stair-hard and block-hard terrains. The blind policy without terrain labels struggles, particularly on complex terrains.

Figure \ref{fig:evaluation} also shows that the teacher and student both have many fewer foot collisions during an episode compared to the blind policies. Both blind policies, especially the baseline without terrain information, experience frequent foot collisions, reflecting a lack of terrain anticipation and proactive control. We see the same trend for the percent of episodes that terminate due to collisions, where the blind policy exhibits the highest failure rates (up to 60\% on block-hard), while both the teacher and student policies demonstrate substantial robustness, maintaining termination rates below 20\% on all terrains.

In terms of energy consumption, measured in kilojoules, the blind policy, trained over a distribution of terrains without explicit terrain input, shows the highest energy cost across all terrains, peaking at over 35 kJ. This is due to the policy adapting to the terrain distribution by converging to a conservative gait strategy characterized by high step heights and exaggerated motions, which anticipates unseen obstacles and results in inefficient and energetically costly locomotion. Both the student and privileged policies achieve significantly lower energy use, indicating smoother and more efficient motions. This is a clear example of how visual information can translate to improved efficiency of a robotic system. 

%Overall, these results highlight the advantages of incorporating vision and privileged training signals. Among the deployable policies, our student policy not only achieves high success rates and safety under difficult terrain conditions but also maintains energy efficiency.

\subsection{Training Time Ablations}

To evaluate the proposed training framework, we compare training durations of the student and teacher across multiple training configurations.  Table~\ref{tab:training_time} shows these comparisons, highlighting the impact of using a pretrained blind policy and visual augmentation for both teacher and student training.

% \begin{table}[!h]
% \centering
% \caption{Comparison of training setups with and without blind policy.}
% \label{tab:training_time}
% \renewcommand{\arraystretch}{1.5}
% % \setlength{\tabcolsep}{6pt}
% \begin{tabular}{|l|p{2.7cm}|p{2.7cm}|}
% \hline
% {Training Phase} & {With Blind Policy}  & {Without Blind Policy} \\
% \hline
% Teacher Training        &   \textbf{152 hours (1.3 Billion)}                &  216 hours (2.0 Billion)                         \\
% \hline
% Student Training        &   \textbf{20 hours (110 Million)}                 &   144 hours (900 Million)                        \\
% \hline
% \end{tabular}
% \end{table}

\begin{table}[!h]
\centering
\caption{Comparison of training setups with and without blind policy.}
\label{tab:training_time}
\renewcommand{\arraystretch}{1.2}
\resizebox{\columnwidth}{!}{%
\begin{tabular}{|l|p{2.7cm}|p{2.7cm}|}
\hline
\textbf{Training Phase} & \textbf{With Blind Policy}  & \textbf{Without Blind Policy} \\
\hline
Teacher Training        & \textbf{152 hours (1.3 Billion)} & 216 hours (2.0 Billion) \\
\hline
Student Training        & \textbf{20 hours (110 Million)} & 144 hours (900 Million) \\
\hline
\begin{tabular}[c]{@{}l@{}}Student Training\\ (No Augmentation)\end{tabular} 
                        & 96 hours (600 Million) & -- \\
\hline
\end{tabular}%
}
\end{table}

For the teacher policy, which utilizes 120 parallel workers, initializing with the pretrained blind policy leads to substantially faster convergence, completing in 152 hours with 1.3 billion samples, compared to 216 hours and 2.0 billion samples without it. These training durations are averaged over 10 independent runs with different random seeds. The results demonstrate that the blind policy serves as a strong prior for exploration, improving sample efficiency and accelerating convergence.

%Together, these results validate our training design: combining a strong blind policy prior with perceptual augmentation dramatically reduces training time, simulation overhead, and rendering costs, facilitating scalable and efficient sim-to-real transfer

The benefits are even more pronounced during student training. When leveraging both the blind policy and our proposed data augmentation strategy, the student policy converges in just 20 hours using 110 million samples. In contrast, removing both components results in significantly slower training, requiring 144 hours and 900 million samples on average across 5 random seeds. To further isolate the utility of the proposed augmentation, we include another configuration where the student is trained using the blind policy but with perceptual augmentation. In this case, we observe that the convergence slows down to 96 hours, requiring about 600 million samples to train. This confirms the importance of augmenting experience buffers and initializing from a pretrained controller to reduce training time, simulation overhead, and rendering cost.

%Overall, these findings validate our design choices of incorporating a pretrained blind controller and data augmentation to accelerates learning and enhances scalability.% enabling efficient sim-to-real transfer with reduced computational resources.

\section{Hardware Deployment}
\label{sec:hardware}
We deployed our proposed system on the bipedal robot Cassie, equipped with four Intel RealSense D455 cameras positioned to provide $360 ^{\circ}$ terrain coverage. For onboard perception, we integrated an NVIDIA Jetson Orin Nano module dedicated to processing visual inputs. The cameras are mounted on custom fixtures and calibrated by aligning depth images to simulated ones under identical robot poses.

Each camera streams depth images at 90 FPS. These are post-processed using hole-filling and distance clipping filters before being passed to the ResNet-based vision encoder running on the Jetson. This module generates a compact vision embedding at 90 Hz, capturing egocentric terrain information around the robot in real time.

The resulting vision embedding is transmitted asynchronously to the robot’s main computer, an Intel NUC, which performs the forward pass of the remainder of the policy network. The policy uses the received vision embedding, proprioceptive input, and control commands to compute the final motor actions. These actions are then sent to the low-level controller for execution.

The end-to-end latency between the vision pipeline and policy inference, including depth streaming, embedding computation, and UDP transmission, is measured to be under 20 milliseconds. To evaluate performance in real-world scenarios, we constructed a test environment using modular wooden blocks of varying heights and sizes, closely mimicking the terrain distributions seen during simulation training (Fig. \ref{fig:lead_fig})

We tested the proposed system on the Cassie bipedal robot across a variety of structured terrains, including single high blocks, stairs, and randomized block configurations with varying dimensions. The system successfully enabled Cassie to traverse these terrains in forward, sideways, and reverse directions. In the forward direction, the robot was able to step up onto blocks as high as 0.5 meters. Due to the robot’s morphological constraints, particularly the limited range of motion and joint configuration. The maximum traversable height was reduced to 0.35 meters for sideways motion and 0.2 meters in the reverse direction. Please refer to the supplementary video for additional real-world deployment demonstrations.
\vspace{-1em}

\section{Summary}

\label{sec:summary}
This work introduces a learning framework for vision-based omnidirectional bipedal locomotion that addresses key challenges in sim-to-real transfer and the high cost of visual rendering for reinforcement learning. The method combines a robust blind controller, a teacher-student distillation approach, and a novel data augmentation strategy to enable scalable and efficient training. We validated the framework in both simulation and real-world settings, demonstrating robust locomotion across diverse terrains.

\bibliographystyle{IEEEtran_link}
\bibliography{references, refs, references_full}

@inproceedings{ross2011,
  title={A reduction of imitation learning and structured prediction to no-regret online learning},
  author={Ross, St{\'e}phane and Gordon, Geoffrey and Bagnell, Drew},
  booktitle={Proceedings of the fourteenth international conference on artificial intelligence and statistics},
  pages={627--635},
  year={2011}
}

@inproceedings{vanMarum2024,
  title={Revisiting reward design and evaluation for robust humanoid standing and walking},
  author={van Marum, Bart and Shrestha, Aayam and Duan, Helei and Dugar, Pranay and Dao, Jeremy and Fern, Alan},
  booktitle={2024 IEEE/RSJ International Conference on Intelligent Robots and Systems (IROS)},
  pages={11256--11263},
  year={2024},
  organization={IEEE}
}

@article{schulman2017proximal,
  title={Proximal policy optimization algorithms},
  author={Schulman, John and Wolski, Filip and Dhariwal, Prafulla and Radford, Alec and Klimov, Oleg},
  journal={arXiv preprint arXiv:1707.06347},
  year={2017}
}

@inproceedings{todorov2012mujoco,
  title={Mujoco: A physics engine for model-based control},
  author={Todorov, Emanuel and Erez, Tom and Tassa, Yuval},
  booktitle={2012 IEEE/RSJ international conference on intelligent robots and systems},
  pages={5026--5033},
  year={2012},
  organization={IEEE}
}

@article{loquercio2022learning,
  title={Learning visual locomotion with cross-modal supervision},
  author={Loquercio, Antonio and Kumar, Ashish and Malik, Jitendra},
  journal={arXiv preprint arXiv:2211.03785},
  year={2022}
}

@article{loquercio_learning_2023,
  title={Learning visual locomotion with cross-modal supervision},
  author={Loquercio, Antonio and Kumar, Ashish and Malik, Jitendra},
  journal={arXiv preprint arXiv:2211.03785},
  year={2022}
}

@article{lee2020learning,
  title={Learning quadrupedal locomotion over challenging terrain},
  author={Lee, Joonho and Hwangbo, Jemin and Wellhausen, Lorenz and Koltun, Vladlen and Hutter, Marco},
  journal={Science robotics},
  volume={5},
  number={47},
  pages={eabc5986},
  year={2020},
  publisher={American Association for the Advancement of Science}
}

@article{jenelten_dtc_2024,
	title = {{DTC}: {Deep} {Tracking} {Control}},
	volume = {9},
	issn = {2470-9476},
	shorttitle = {{DTC}},
	url = {http://arxiv.org/abs/2309.15462},
	doi = {10.1126/scirobotics.adh5401},
	number = {86},
	urldate = {2024-03-06},
	journal = {Science Robotics},
	author = {Jenelten, Fabian and He, Junzhe and Farshidian, Farbod and Hutter, Marco},
	month = jan,
	year = {2024},
	note = {arXiv:2309.15462 [cs, eess]},
	pages = {eadh5401},
}

@misc{bewley_learning_2018,
	title = {Learning to {Drive} from {Simulation} without {Real} {World} {Labels}},
	url = {http://arxiv.org/abs/1812.03823},
	doi = {10.48550/arXiv.1812.03823},
	urldate = {2025-05-16},
	publisher = {arXiv},
	author = {Bewley, Alex and Rigley, Jessica and Liu, Yuxuan and Hawke, Jeffrey and Shen, Richard and Lam, Vinh-Dieu and Kendall, Alex},
	month = dec,
	year = {2018},
	note = {arXiv:1812.03823 [cs]},
}

@misc{bansal_chauffeurnet_2018,
	title = {{ChauffeurNet}: {Learning} to {Drive} by {Imitating} the {Best} and {Synthesizing} the {Worst}},
	shorttitle = {{ChauffeurNet}},
	url = {http://arxiv.org/abs/1812.03079},
	doi = {10.48550/arXiv.1812.03079},
	urldate = {2025-05-16},
	publisher = {arXiv},
	author = {Bansal, Mayank and Krizhevsky, Alex and Ogale, Abhijit},
	month = dec,
	year = {2018},
	note = {arXiv:1812.03079 [cs]},
}

@article{wang_cts_2024,
	title = {{CTS}: {Concurrent} {Teacher}-{Student} {Reinforcement} {Learning} for {Legged} {Locomotion}},
	volume = {9},
	issn = {2377-3766},
	shorttitle = {{CTS}},
	url = {https://ieeexplore.ieee.org/document/10670293/},
	doi = {10.1109/LRA.2024.3457379},
	number = {11},
	urldate = {2025-05-16},
	journal = {IEEE Robotics and Automation Letters},
	author = {Wang, Hongxi and Luo, Haoxiang and Zhang, Wei and Chen, Hua},
	month = nov,
	year = {2024},
	pages = {9191--9198},
}

@misc{hoeller_neural_2022,
	title = {Neural {Scene} {Representation} for {Locomotion} on {Structured} {Terrain}},
	url = {http://arxiv.org/abs/2206.08077},
	doi = {10.48550/arXiv.2206.08077},
	urldate = {2025-05-15},
	publisher = {arXiv},
	author = {Hoeller, David and Rudin, Nikita and Choy, Christopher and Anandkumar, Animashree and Hutter, Marco},
	month = jun,
	year = {2022},
	note = {arXiv:2206.08077 [cs]},
}

@misc{miki_elevation_2022,
	title = {Elevation {Mapping} for {Locomotion} and {Navigation} using {GPU}},
	url = {http://arxiv.org/abs/2204.12876},
	doi = {10.48550/arXiv.2204.12876},
	urldate = {2025-05-15},
	publisher = {arXiv},
	author = {Miki, Takahiro and Wellhausen, Lorenz and Grandia, Ruben and Jenelten, Fabian and Homberger, Timon and Hutter, Marco},
	month = apr,
	year = {2022},
	note = {arXiv:2204.12876 [cs]},
}

@misc{chen_vmts_2025,
	title = {{VMTS}: {Vision}-{Assisted} {Teacher}-{Student} {Reinforcement} {Learning} for {Multi}-{Terrain} {Locomotion} in {Bipedal} {Robots}},
	shorttitle = {{VMTS}},
	url = {http://arxiv.org/abs/2503.07049},
	doi = {10.48550/arXiv.2503.07049},
	urldate = {2025-05-15},
	publisher = {arXiv},
	author = {Chen, Fu and Wan, Rui and Liu, Peidong and Zheng, Nanxing and Zhou, Bo},
	month = mar,
	year = {2025},
	note = {arXiv:2503.07049 [cs]
version: 1},
}

@misc{radosavovic_learning_2024,
	title = {Learning {Humanoid} {Locomotion} over {Challenging} {Terrain}},
	url = {http://arxiv.org/abs/2410.03654},
	doi = {10.48550/arXiv.2410.03654},
	urldate = {2025-05-15},
	publisher = {arXiv},
	author = {Radosavovic, Ilija and Kamat, Sarthak and Darrell, Trevor and Malik, Jitendra},
	month = oct,
	year = {2024},
	note = {arXiv:2410.03654 [cs]},
}

@misc{wu_learn_2025,
	title = {Learn to {Teach}: {Sample}-{Efficient} {Privileged} {Learning} for {Humanoid} {Locomotion} over {Diverse} {Terrains}},
	shorttitle = {Learn to {Teach}},
	url = {http://arxiv.org/abs/2402.06783},
	doi = {10.48550/arXiv.2402.06783},
	urldate = {2025-05-15},
	publisher = {arXiv},
	author = {Wu, Feiyang and Nal, Xavier and Jang, Jaehwi and Zhu, Wei and Gu, Zhaoyuan and Wu, Anqi and Zhao, Ye},
	month = mar,
	year = {2025},
	note = {arXiv:2402.06783 [cs]},
}

@inproceedings{siekmann_blind_2021,
	title = {Blind {Bipedal} {Stair} {Traversal} via {Sim}-to-{Real} {Reinforcement} {Learning}},
	isbn = {978-0-9923747-7-8},
	url = {http://www.roboticsproceedings.org/rss17/p061.pdf},
	doi = {10.15607/RSS.2021.XVII.061},
	language = {en},
	urldate = {2025-05-15},
	booktitle = {Robotics: {Science} and {Systems} {XVII}},
	publisher = {Robotics: Science and Systems Foundation},
	author = {Siekmann, Jonah and Green, Kevin and Warila, John and Fern, Alan and Hurst, Jonathan},
	month = jul,
	year = {2021},
}

@inproceedings{zhang_resilient_2024,
	title = {Resilient {Legged} {Local} {Navigation}: {Learning} to {Traverse} with {Compromised} {Perception} {End}-to-{End}},
	shorttitle = {Resilient {Legged} {Local} {Navigation}},
	url = {https://ieeexplore.ieee.org/document/10611254/?arnumber=10611254},
	doi = {10.1109/ICRA57147.2024.10611254},
	urldate = {2025-03-18},
	booktitle = {2024 {IEEE} {International} {Conference} on {Robotics} and {Automation} ({ICRA})},
	author = {Zhang, Chong and Jin, Jin and Frey, Jonas and Rudin, Nikita and Mattamala, Matías and Cadena, Cesar and Hutter, Marco},
	month = may,
	year = {2024},
	pages = {34--41},
}

@inproceedings{duan_learning_2024,
	title = {Learning {Vision}-{Based} {Bipedal} {Locomotion} for {Challenging} {Terrain}},
	url = {https://ieeexplore.ieee.org/document/10611621/?arnumber=10611621},
	doi = {10.1109/ICRA57147.2024.10611621},
	urldate = {2025-03-18},
	booktitle = {2024 {IEEE} {International} {Conference} on {Robotics} and {Automation} ({ICRA})},
	author = {Duan, Helei and Pandit, Bikram and Gadde, Mohitvishnu S. and Van Marum, Bart and Dao, Jeremy and Kim, Chanho and Fern, Alan},
	month = may,
	year = {2024},
	pages = {56--62},
}

@inproceedings{agarwal_legged_2023,
	title = {Legged {Locomotion} in {Challenging} {Terrains} using {Egocentric} {Vision}},
	url = {https://proceedings.mlr.press/v205/agarwal23a.html},
	language = {en},
	urldate = {2025-03-18},
	booktitle = {Proceedings of {The} 6th {Conference} on {Robot} {Learning}},
	publisher = {PMLR},
	author = {Agarwal, Ananye and Kumar, Ashish and Malik, Jitendra and Pathak, Deepak},
	month = mar,
	year = {2023},
	note = {ISSN: 2640-3498},
	pages = {403--415},
}

@misc{zhuang_humanoid_2024,
	title = {Humanoid {Parkour} {Learning}},
	url = {http://arxiv.org/abs/2406.10759},
	doi = {10.48550/arXiv.2406.10759},
	urldate = {2025-03-18},
	publisher = {arXiv},
	author = {Zhuang, Ziwen and Yao, Shenzhe and Zhao, Hang},
	month = sep,
	year = {2024},
	note = {arXiv:2406.10759 [cs]},
}

@inproceedings{hu_privileged_2023,
	title = {Privileged {Sensing} {Scaffolds} {Reinforcement} {Learning}},
	url = {https://openreview.net/forum?id=EpVe8jAjdx},
	language = {en},
	urldate = {2024-04-26},
	booktitle = {Proceedings of the 7th Conference on Robot Learning (CoRL)},
	author = {Hu, Edward S. and Springer, James and Rybkin, Oleh and Jayaraman, Dinesh},
	month = oct,
	year = {2023},
}

@inproceedings{chen_learning_2020,
	title = {Learning by {Cheating}},
	url = {https://proceedings.mlr.press/v100/chen20a.html},
	language = {en},
	urldate = {2024-04-30},
	booktitle = {Proceedings of the {Conference} on {Robot} {Learning}},
	publisher = {PMLR},
	author = {Chen, Dian and Zhou, Brady and Koltun, Vladlen and Krähenbühl, Philipp},
	month = may,
	year = {2020},
	pages = {66--75},
}

@misc{pinto_asymmetric_2017,
	title = {Asymmetric {Actor} {Critic} for {Image}-{Based} {Robot} {Learning}},
	url = {http://arxiv.org/abs/1710.06542},
	doi = {10.48550/arXiv.1710.06542},
	urldate = {2024-04-29},
	publisher = {arXiv},
	author = {Pinto, Lerrel and Andrychowicz, Marcin and Welinder, Peter and Zaremba, Wojciech and Abbeel, Pieter},
	month = oct,
	year = {2017},
	note = {arXiv:1710.06542 [cs]},
}

@inproceedings{chen_system_2022,
	title = {A {System} for {General} {In}-{Hand} {Object} {Re}-{Orientation}},
	url = {https://proceedings.mlr.press/v164/chen22a.html},
	language = {en},
	urldate = {2024-04-29},
	booktitle = {Proceedings of the 5th {Conference} on {Robot} {Learning}},
	publisher = {PMLR},
	author = {Chen, Tao and Xu, Jie and Agrawal, Pulkit},
	month = jan,
	year = {2022},
	pages = {297--307},
}

@misc{yamada_twist_2023,
	title = {{TWIST}: {Teacher}-{Student} {World} {Model} {Distillation} for {Efficient} {Sim}-to-{Real} {Transfer}},
	shorttitle = {{TWIST}},
	url = {http://arxiv.org/abs/2311.03622},
	doi = {10.48550/arXiv.2311.03622},
	urldate = {2024-04-26},
	publisher = {arXiv},
	author = {Yamada, Jun and Rigter, Marc and Collins, Jack and Posner, Ingmar},
	month = nov,
	year = {2023},
	note = {arXiv:2311.03622 [cs]},
}

@inproceedings{duan_learning_2022,
	title = {Learning {Dynamic} {Bipedal} {Walking} {Across} {Stepping} {Stones}},
	url = {https://ieeexplore.ieee.org/document/9981884/?arnumber=9981884},
	doi = {10.1109/IROS47612.2022.9981884},
	urldate = {2025-03-18},
	booktitle = {2022 {IEEE}/{RSJ} {International} {Conference} on {Intelligent} {Robots} and {Systems} ({IROS})},
	author = {Duan, Helei and Malik, Ashish and Gadde, Mohitvishnu S. and Dao, Jeremy and Fern, Alan and Hurst, Jonathan},
	month = oct,
	year = {2022},
	note = {ISSN: 2153-0866},
	pages = {6746--6752},
}

@misc{margolis_learning_2021,
	title = {Learning to {Jump} from {Pixels}},
	url = {http://arxiv.org/abs/2110.15344},
	doi = {10.48550/arXiv.2110.15344},
	urldate = {2025-01-09},
	publisher = {arXiv},
	author = {Margolis, Gabriel B. and Chen, Tao and Paigwar, Kartik and Fu, Xiang and Kim, Donghyun and Kim, Sangbae and Agrawal, Pulkit},
	month = oct,
	year = {2021},
	note = {arXiv:2110.15344 [cs]},
}

@article{miki_learning_2022,
	title = {Learning robust perceptive locomotion for quadrupedal robots in the wild},
	volume = {7},
	url = {https://www.science.org/doi/abs/10.1126/scirobotics.abk2822},
	doi = {10.1126/scirobotics.abk2822},
	number = {62},
	urldate = {2025-01-09},
	journal = {Science Robotics},
	author = {Miki, Takahiro and Lee, Joonho and Hwangbo, Jemin and Wellhausen, Lorenz and Koltun, Vladlen and Hutter, Marco},
	month = jan,
	year = {2022},
	note = {Publisher: American Association for the Advancement of Science},
	pages = {eabk2822},
}

@article{yu_visual-locomotion_nodate,
	title = {Visual-{Locomotion}: {Learning} to {Walk} on {Complex} {Terrains} with {Vision}},
	language = {en},
	journal = {arXiv preprint arXiv:2107.07022},
	author = {Yu, Wenhao and Jain, Deepali and Escontrela, Alejandro and Iscen, Atil and Xu, Peng and Coumans, Erwin and Ha, Sehoon and Tan, Jie and Zhang, Tingnan},
	year = {2021},
}

@phdthesis{fankhauser_perceptive_2018,
	type = {Doctoral {Thesis}},
	title = {Perceptive {Locomotion} for {Legged} {Robots} in {Rough} {Terrain}},
	copyright = {http://rightsstatements.org/page/InC-NC/1.0/},
	url = {https://www.research-collection.ethz.ch/handle/20.500.11850/284254},
	language = {en},
	urldate = {2025-01-09},
	school = {ETH Zurich},
	author = {Fankhauser, Péter},
	year = {2018},
	doi = {10.3929/ethz-b-000284254},
	note = {Accepted: 2018-08-24T07:58:52Z},
}

@misc{zhuang_robot_2023,
	title = {Robot {Parkour} {Learning}},
	url = {http://arxiv.org/abs/2309.05665},
	doi = {10.48550/arXiv.2309.05665},
	urldate = {2025-01-09},
	publisher = {arXiv},
	author = {Zhuang, Ziwen and Fu, Zipeng and Wang, Jianren and Atkeson, Christopher and Schwertfeger, Soeren and Finn, Chelsea and Zhao, Hang},
	month = sep,
	year = {2023},
	note = {arXiv:2309.05665 [cs]},
}

@misc{cheng_extreme_2023,
	title = {Extreme {Parkour} with {Legged} {Robots}},
	url = {http://arxiv.org/abs/2309.14341},
	doi = {10.48550/arXiv.2309.14341},
	urldate = {2025-01-09},
	publisher = {arXiv},
	author = {Cheng, Xuxin and Shi, Kexin and Agarwal, Ananye and Pathak, Deepak},
	month = sep,
	year = {2023},
	note = {arXiv:2309.14341 [cs]},
}

\end{document}